\documentclass{ifacconf}

\usepackage{graphicx}      
\usepackage{natbib}        
\usepackage{amsmath}
\usepackage{amssymb}
\usepackage{booktabs}
\begin{document}
\begin{frontmatter}

\title{Conditional Invertible Neural Networks for Data-Driven UAV Control: A 2-D Proof of Concept} 

\author[First]{Christian Wittke} 
\author[Second]{Stephan Myschik} 
\author[First]{Oliver Niggemann}

\address[First]{Computer Science in Mechanical Engineering, Helmut Schmidt University, Hamburg, Germany.\newline(e-mail: christian.wittke;oliver.niggemann@hsu-hh.de)}
\address[Second]{Institute for Aeronautical Engineering, University of the Bundeswehr Munich, Bavaria, Germany (e-mail: stephan.myschik@unibw.de)}

\begin{abstract}
We investigate conditional invertible neural networks (cINNs) as
probabilistic inverse-dynamics models for multirotor control. For a
planar X8 coaxial multicopter, we learn $p(u \mid s_t, c_t)$ from
an incremental nonlinear dynamic inversion (INDI) teacher using
rational-quadratic spline coupling and invertible linear mixing.
Open-loop reproduction reaches $R^2 = 0.944$, mean CRPS $0.0915$,
and log-probability--error correlation $\rho = -0.60$. Over $15$
closed-loop scenarios, position RMSE matches INDI ($9.7$ vs.\
$9.5\,\mathrm{m}$) with $47\,\%$ tracking acceptably; failures
separate into attitude divergence under aggressive steps and phase
lag under high-frequency references, isolating command bandwidth
and data coverage as dominant failure mechanisms.
\end{abstract}

\begin{keyword}
Unmanned aerial vehicles, Flight control, Learning and adaptive systems, Neural networks, Probabilistic methods
\end{keyword}

\end{frontmatter}
\begingroup
  \renewcommand{\thefootnote}{}
  \footnotetext[0]{%
    \textcopyright{} 2026 the authors. This work has been accepted to IFAC
    for publication under a Creative Commons Licence CC-BY-NC-ND.%
  }
\endgroup
\enlargethispage{3\baselineskip}

\section{Introduction}

\label{sec:intro}

Multirotor control has matured to the point where time-optimal
trajectories are flown by both model-based and learned controllers,
yet a gap remains between the two paradigms. Classical linear
control methods such as PID and model predictive control (MPC), as
well as controllers based on (incremental) nonlinear dynamic
inversion (NDI/INDI), are predictable and certifiable but rely on
accurate system models. Data-driven methods generalise from
experience but return point estimates without any associated
confidence. Neither is sufficient in safety-relevant deployments,
where the operator needs both adaptability to unmodelled effects
and a quantitative signal that flags when the controller leaves
its trusted regime.

This paper treats multirotor control as a probabilistic inverse
dynamics problem. Given the current state $s_t$ and a tracking
command $c_t$, we model the full conditional distribution
\begin{equation}
p(u \mid s_t, c_t)
\label{eq:framing}
\end{equation}
over the motor command $u \in \mathbb{R}^{n_u}$ instead of
predicting a single value. The mode of $p(u\mid s_t, c_t)$ serves
as the deterministic control output; its dispersion provides a
sample-wise uncertainty estimate that contracts where training data
is dense and spreads where it is sparse, the behaviour conditional
flows exhibit on other inverse problems
\citep{Ardizzone2019Analyzing,denker2021conditional}. The
construction mirrors INDI, which builds its control update from
the current state and a desired acceleration encoded implicitly in
$c_t$ through the velocity command and tracking errors.

To represent~\eqref{eq:framing} we use a conditional invertible
neural network (cINN) built from normalising flows
\citep{dinh2017density,Kingma2018Glow,durkan2019neural,Ardizzone2019Analyzing}.
cINNs combine three properties relevant here: a bijection between
data and a tractable base distribution, exact likelihood evaluation
via the change-of-variables formula, and efficient sampling. We use
rational-quadratic spline coupling \citep{durkan2019neural} for the
motor space, activation normalisation \citep{Kingma2018Glow} for
stable optimisation, and learned invertible linear mixing for
cross-channel coupling, with sine--cosine encoding of attitude
angles in the conditioning network.

The testbed is a planar X8 coaxial multicopter restricted to the
$XZ$-plane, with eight motor inputs and reference velocity commands
up to $\pm 2\,\mathrm{m/s}$ in $x$ and $z$. The simulation
framework extends a Simulink model used in prior
sparse-identification work on the same vehicle
\citep{tappe2025qualitative,kelm2023sindyc}. An INDI teacher
supplies both the training data and the closed-loop baseline. The
trained cINN reproduces the teacher across all eight motor
channels at $R^2 = 0.944$ and matches the INDI baseline in mean
position RMSE (Root Mean Square Error) over $15$ closed-loop scenarios. Two failure modes
emerge---attitude divergence under aggressive single-step commands
and phase lag under high-frequency references---both with concrete
physical origins.
\newline
The main contributions are:
\begin{itemize}
\item a probabilistic inverse-dynamics formulation
      $p(u \mid s_t, c_t)$ that conditions on operationally
      available signals rather than the unobservable future state
      $s_{t+1}$;
\item a cINN architecture combining rational-quadratic spline
      coupling, activation normalisation, and learned invertible
      linear mixing, with sine--cosine attitude encoding;
\item an open- and closed-loop evaluation on a 2-D X8 with an INDI
      teacher as both data source and baseline, isolating command
      bandwidth and training-data coverage as the dominant failure
      mechanisms.
\end{itemize}

Section~\ref{sec:sota} reviews related work;
Section~\ref{sec:solution} formalises the problem and the cINN
architecture; Section~\ref{sec:methodology} the experimental setup;
Section~\ref{sec:results} the results;
Section~\ref{sec:conclusion} concludes with future work.

\section{State of the Art}
%
%
%

\label{sec:sota}

\subsection{Multirotor Control}

Cascaded proportional-integral-derivative (PID) loops remain the
dominant control architecture for commercial multirotors thanks to
their transparency and well-understood tuning procedures
\citep{lopez2023pid,Chao2010Autopilots}. Linear control laws,
however, struggle with the coupling and nonlinearities of
aggressive flight, which has motivated nonlinear extensions: model
predictive control (MPC) optimizes over a receding horizon and has
enabled near time-optimal racing performance
\citep{romero2022model,foehn2021time}, while incremental nonlinear
dynamic inversion (INDI) treats the inverse of the rotational
dynamics explicitly and has become the de facto standard for
aggressive flight on resource-constrained platforms
\citep{smeur2016adaptive}. Both approaches require accurate parameter knowledge and provide no
native measure of prediction confidence---a gap that becomes critical
outside the trusted operating regime.

\subsection{Learning-Based and Imitation Approaches}

Supervised imitation of expert controllers is a well-established
paradigm for synthesising fast neural policies
\citep{dierks2009neural,loquercio2020deep}. Reinforcement learning
removes the need for an explicit teacher and has produced
champion-level drone racing controllers
\citep{Kaufmann2023Champion}, with comparable performance limits
reported for both RL and optimal control under appropriate
conditions \citep{song2023reaching}. Hybrid strategies combine
learned components with structural priors, including
physics-informed neural networks \citep{cheng2024physics} and
qualitative system models \citep{tappe2025qualitative}.
Almost all of these controllers emit deterministic motor commands;
where uncertainty is reported, it relies on Monte Carlo dropout or
deep ensembles, which lack a principled likelihood interpretation.
We position the present work as supervised imitation of an INDI
teacher with the network's probabilistic output replacing such
ad-hoc estimators.

\subsection{Conditional Normalizing Flows for Inverse Problems}

Normalizing flows model complex densities as compositions of
invertible transformations, enabling exact likelihood evaluation
through the change-of-variables formula
\citep{dinh2017density,Kingma2018Glow}. Affine coupling layers and
learned $1\!\times\!1$ mixings provide tractable Jacobians, while
rational-quadratic spline flows substantially increase
expressiveness without sacrificing analytic invertibility
\citep{durkan2019neural}. Conditional extensions \citep{Ardizzone2019Analyzing} model
$p(x \mid c)$ and have been deployed across inverse problems in
medical imaging~\citep{denker2021conditional,nolke2021invertible},
materials design~\citep{kumar2021inverse}, particle
physics~\citep{heredge2024generative}, industrial
vision~\citep{PotatoGlow2024}, and continuum-robot inverse
kinematics~\citep{rao2023inverse}. The probabilistic
inverse-dynamics setting on aerial vehicles has, to our knowledge,
not been explored.

\section{Solution}

%
%
%
\label{sec:solution}

\subsection{Problem Formulation}

The inverse-dynamics problem asks for the motor command $u \in
\mathbb{R}^{n_u}$ that drive the vehicle from $s_t \in
\mathbb{R}^{n_s}$ along an external reference. Conditioning on the
desired next state $s_{t+1}$ is unavailable in closed loop, since
$s_{t+1}$ is not observable when the command must be issued. We
condition instead on the signals an outer loop produces in practice:
the velocity command $v_t^{\mathrm{cmd}}$ and the position and
velocity tracking errors $e_t^{\mathrm{pos}}$, $e_t^{\mathrm{vel}}$,
collected in
\begin{equation}
c_t = \bigl[v_t^{\mathrm{cmd}};\; e_t^{\mathrm{pos}};\;
            e_t^{\mathrm{vel}}\bigr] \in \mathbb{R}^{n_c}
\end{equation}
This is the same information INDI uses to construct its control
update. We learn the full conditional density $p(u \mid s_t, c_t)$,
which captures the teacher's response together with the dispersion
induced by limited training coverage.

\subsection{System Model}

The vehicle is an X8 coaxial multicopter with eight individually
controlled rotors in four counter-rotating coaxial pairs. The plant
runs in Simulink, exposed to Python as a precompiled C library
through the RMT-CopterGym environment of
\citet{tappe2025qualitative}. The motor command is $u = [u_1,
\ldots, u_8]^\top \in \mathbb{R}^8$, and the state aggregates
position, linear velocity, attitude, and body angular rates:
$s = [\mathbf{p}; \mathbf{v}; \boldsymbol{\Theta}; \boldsymbol{\omega}]
\in \mathbb{R}^{12}$ with $\boldsymbol{\Theta} = [\phi, \theta,
\psi]^\top$. For the proof of concept the simulation is restricted
to the $XZ$-plane: lateral position and velocity, roll, yaw, and the
corresponding angular rates remain identically zero. The full $12$-D
state is retained so the architecture extends to 3-D flight without
modification.

\subsection{Conditional Invertible Neural Network}

The cINN learns a parametric bijection
$f_\theta(\,\cdot\,;\,h)\colon \mathbb{R}^{n_u} \to
\mathbb{R}^{n_u}$, $z = f_\theta(u;\,h)$, that maps motor commands
to a latent vector $z \sim p_z = \mathcal{N}(0, I_{n_u})$. A learned
encoder $g_\theta$ produces the conditioning vector $h \in
\mathbb{R}^{n_h}$ from $(s_t, c_t)$. The change-of-variables formula
gives the conditional density of $u$ in closed form:
\begin{equation}
\log p\bigl(u \mid s_t, c_t\bigr)
= \log p_z\bigl(f_\theta(u;\,h)\bigr)
  + \log \biggl|\det \frac{\partial f_\theta}{\partial u}\biggr|
\label{eq:logprob}
\end{equation}
A standard normal base is sufficient: spline-based bijections are
universal density approximators \citep{durkan2019neural}.

\subsection{Conditioning Encoder}

Three of the twelve state components are angles
($\boldsymbol{\Theta} = [\phi,\theta,\psi]^\top$) and contain a
$2\pi$-discontinuity that an MLP cannot fit smoothly. We replace
them by their sine--cosine encoding, yielding a $15$-D state and a
$24$-D encoder input together with the $9$-D command. The encoder
$g_\theta$ is a three-layer MLP with hidden dimension $256$ and SiLU
activations, producing $h \in \mathbb{R}^{256}$ which conditions
every flow block.

\subsection{Flow Architecture}

\begin{figure*}
\centering
\includegraphics[width=0.85\textwidth]{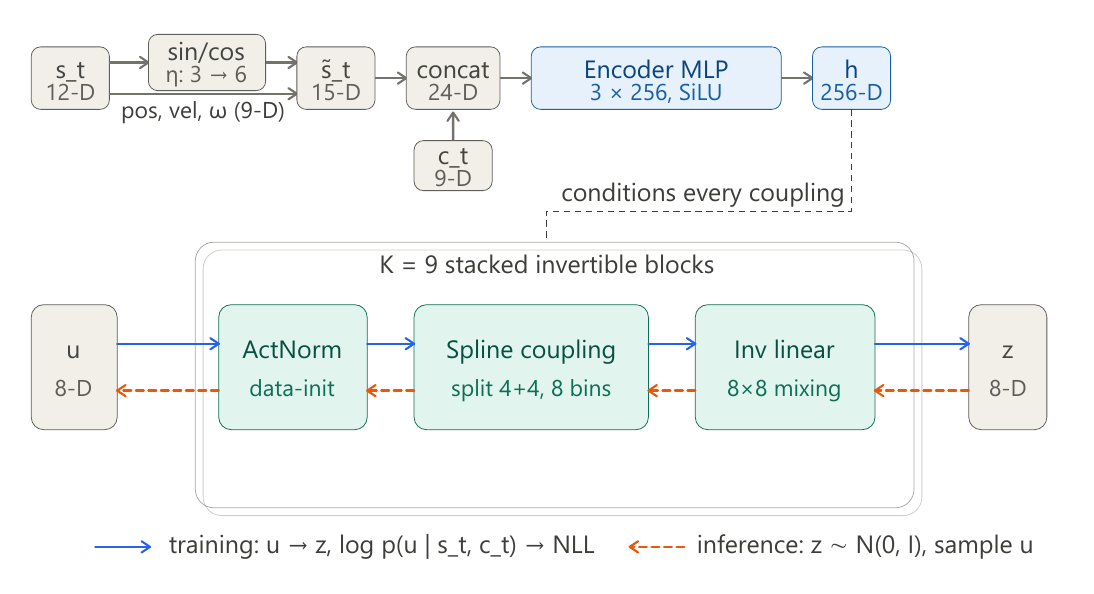}
\caption{cINN architecture. The encoder maps the augmented state
$\tilde s_t$ and command $c_t$ to a context $h$, which conditions
each of $K=9$ flow blocks. Each block consists of activation
normalisation, rational-quadratic spline coupling, and a learned
invertible linear map; the base distribution is
$\mathcal{N}(0, I_8)$.}
\label{fig:architecture}
\end{figure*}

The flow $f_\theta$ stacks $K = 9$ blocks, each combining three
invertible sub-layers (Fig.~\ref{fig:architecture}). Activation
normalisation \citep{Kingma2018Glow} applies an affine per-channel
scale with parameters initialised from the first mini-batch. A
rational-quadratic spline coupling layer \citep{durkan2019neural}
splits the $8$-D input into two halves; one half is transformed
componentwise by a monotonic piecewise spline ($K_s = 8$ bins on
$[-B, B]$) whose parameters are predicted from the other half and
$h$. Spline coupling outperforms affine coupling
\citep{dinh2017density} here because motor allocation in a coaxial
X8 is strongly nonlinear. A learned $8\!\times\!8$ linear map
\citep{Kingma2018Glow}, initialised orthogonal, mixes across
channels and compensates for the static coupling split. The full
network has about $1.5\!\times\!10^{6}$ trainable parameters.

\subsection{Training and Inference}

Training minimises the negative log-likelihood of $u$ over the
dataset $\mathcal{D}$, with $\log p(u \mid s_t, c_t)$ evaluated
via~\eqref{eq:logprob}. At inference, the deterministic control
output is the mode $\hat u_t = f_\theta^{-1}(0;\,h)$; for
uncertainty quantification we draw $N = 20$ latent samples through
the inverse flow and report the empirical mean and componentwise
standard deviation.

\section{Methodology}

%
%
%
%

\label{sec:methodology}

\subsection{Simulation and Data}

All experiments run inside the RMT-CopterGym environment of
\citet{tappe2025qualitative} at a fixed control rate of
$100\,$Hz. The simulation is constrained to the $XZ$-plane and
uses the full $12$-D simulator state as conditioning input.
Training data is generated by an INDI teacher
\citep{smeur2016adaptive}; INDI uses the measured (here: simulated)
angular acceleration to invert the rotational dynamics
incrementally, yielding stable high-bandwidth velocity tracking
without requiring a precise inertia model. The same INDI controller
also serves as the closed-loop baseline.

Reference signals are piecewise-constant velocity commands
$v_t^{\mathrm{cmd}} \in \mathbb{R}^3$ with non-zero entries only
in $x$ and $z$, step amplitudes drawn uniformly from
$\pm 2\,\mathrm{m/s}$, switching times from a coarse
$1$--$3\,\mathrm{s}$ grid. Each scenario runs for $12.5\,\mathrm{s}$
($1\,252$ samples); the full dataset comprises approximately
$\text{1000}$ scenarios totalling ${\sim}1.25\!\times\!10^{6}$
tuples $(s_t, c_t, u_t)$, with $c_t$ constructed online from the
instantaneous tracking errors.

States, commands, and controls are standardised channel-wise using
training-split statistics, frozen for all subsequent use.
Trajectories are split scenario-wise into training, validation, and
test sets ($75/15/10\,\%$), so every time-step belongs to exactly
one split.

\subsection{Network and Training}

Table~\ref{tab:config} summarises the architecture and the training
hyperparameters. The model is optimised with AdamW on the NLL
objective, with a linear warmup over $5$ epochs and
cosine annealing of the learning rate to $10^{-6}$. An exponential
moving average of the weights is used at evaluation. We use early
stopping on the validation NLL: training is terminated when the
validation loss has not improved for $15$ consecutive epochs. With
this rule, training stops around epoch $75$; total wall-clock time
on a single NVIDIA RTX~$3090$ is $2\,$h~$52\,$min.

\begin{table}[h]
\centering
\caption{cINN architecture and training configuration.}
\label{tab:config}
\small
\begin{tabular}{ll}
\toprule
State / command / control dim & $12$ / $9$ / $8$ \\
Encoder & $3 \times 256$, SiLU \\
Context dim $n_h$ & $256$ \\
Flow blocks $\times$ layers $K$ & $3 \times 3 = 9$ \\
Spline bins $K_s$, tail bound $B$ & $8$, $5.0$ \\
Trainable parameters & ${\sim}1.5\!\times\!10^{6}$ \\
\midrule
Optimiser & AdamW, lr $3\!\times\!10^{-4}$ \\
Batch size, weight decay & $512$, $10^{-5}$ \\
Schedule & cosine, $5$-epoch warmup \\
EMA decay, gradient clip & $0.999$, $1.0$ \\
Early-stopping patience & $15$ epochs \\
\bottomrule
\end{tabular}
\end{table}

\subsection{Evaluation}

We evaluate the trained cINN in two complementary regimes. In
\emph{open-loop} evaluation, the network sees ground-truth
state--command pairs $(s_t, c_t)$ from the held-out test set and we
compare its predicted motor command against the ground-truth INDI
command at the same time-step; the simulator is not in the loop.
In \emph{closed-loop} evaluation, the network replaces INDI in the
inner control loop: at every $10\,\mathrm{ms}$ step the simulator
state is read out, $c_t$ is built from the current velocity
command and tracking errors, and the predicted motor command is
written back to the plant. The two regimes thus probe different
properties: open-loop quantifies one-step prediction accuracy and
calibration; closed-loop reveals whether the small per-step errors
compound into stable or unstable trajectories.

Open-loop evaluation on the $125\,100$ held-out test samples
reports the coefficient of determination $R^2$, root-mean-square
error (RMSE), and mean absolute error (MAE) of the mode prediction
$\hat u_t = f_\theta^{-1}(0;\,h)$, both per channel and aggregated.
Probabilistic quality is assessed via the continuous ranked
probability score (CRPS) computed from $N = 20$ samples per test
point and the empirical coverage at $68\,\%$ and $95\,\%$ nominal
levels; we apply a single post-hoc temperature $T$ minimising the
validation NLL.

Closed-loop evaluation uses $15$ held-out scenarios disjoint from
training. For each, we run the simulator with INDI and with the
cINN replacing INDI in the inner loop, and log position RMSE,
final position error, peak pitch excursion, and episode duration
up to either nominal termination or the simulator's attitude
safety break. A single inference takes $47.5\,\mathrm{ms}$ on the
RTX~$3090$, longer than the $10\,\mathrm{ms}$ control cycle, so
the evaluation uses the simulator's time abstraction to step the
cINN deterministically once per control step.

\section{Results}
\label{sec:results}
%
%
%
%

\subsection{Open-Loop Reproduction}

\begin{figure}
\centering
\includegraphics[width=\columnwidth]{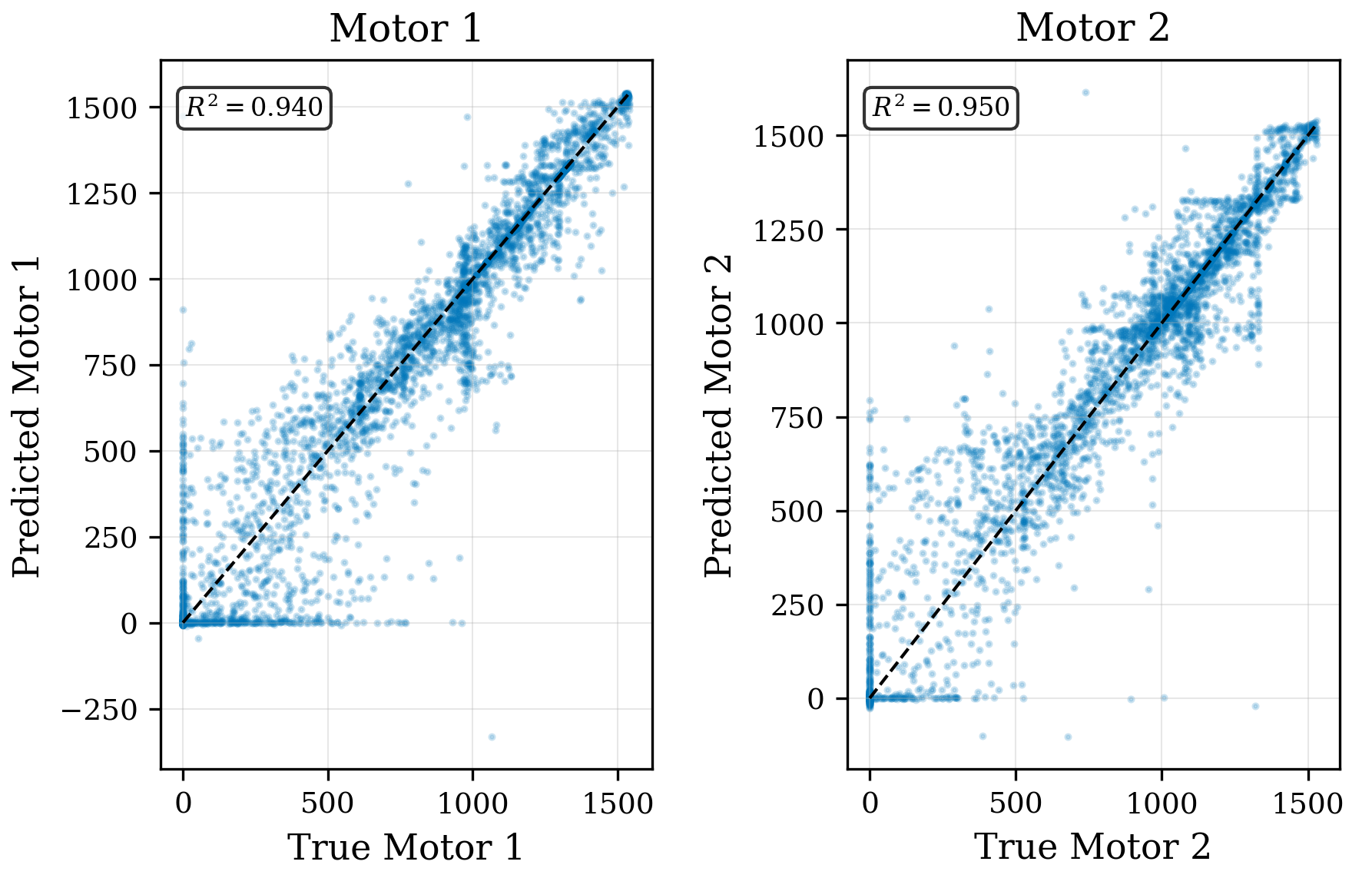}
\caption{Predicted vs.\ ground-truth motor commands for Motor~$1$
($R^2 = 0.940$, per-channel worst) and Motor~$2$ ($R^2 = 0.950$,
per-channel best); other motors fall between these two.}
\label{fig:scatter}
\end{figure}

Across $125\,100$ held-out test samples the cINN reproduces the
INDI teacher with mean $R^2 = 0.944$, RMSE $121.0\,\mathrm{rad/s}$
($7.8\,\%$ of the operating range), and MAE $62.0\,\mathrm{rad/s}$.
Per-channel $R^2$ stays within $0.940$--$0.950$; the small
asymmetry between Motors~$1,7$ and Motors~$2,8$ reflects the
diagonal pairing of coaxial rotors in the $XZ$-plane. Aggregate
metrics are in Table~\ref{tab:performance};
Fig.~\ref{fig:scatter} shows the prediction vs.\ ground-truth
relationship for the two motors that bracket the per-channel range.

\begin{table}[h]
\centering
\caption{cINN open-loop performance over $8$ motor channels and
$125\,100$ test samples (mean and range across motors). Inference:
$47.45\,\mathrm{ms}$ per sample on RTX~$3090$.}
\label{tab:performance}
\small
\begin{tabular}{lcc}
\toprule
\textbf{Metric} & \textbf{Mean} & \textbf{Range} \\
\midrule
RMSE [rad/s]      & $121.0$  & $116.8$--$127.6$ \\
RMSE [\% range]   & $7.8$    & $7.2$--$8.3$ \\
MAE [rad/s]       & $62.0$   & $58.2$--$64.9$ \\
$R^2$             & $0.944$  & $0.940$--$0.950$ \\
CRPS              & $0.0915$ & $0.0858$--$0.0974$ \\
Coverage $68\,\%$ & $57.7\,\%$ & $56.0$--$59.0\,\%$ \\
Coverage $95\,\%$ & $73.9\,\%$ & $72.1$--$75.6\,\%$ \\
NLL               & $34.76$  & $\pm 15.03$ \\
\bottomrule
\end{tabular}
\end{table}

\begin{figure}
\centering
\includegraphics[width=\columnwidth]{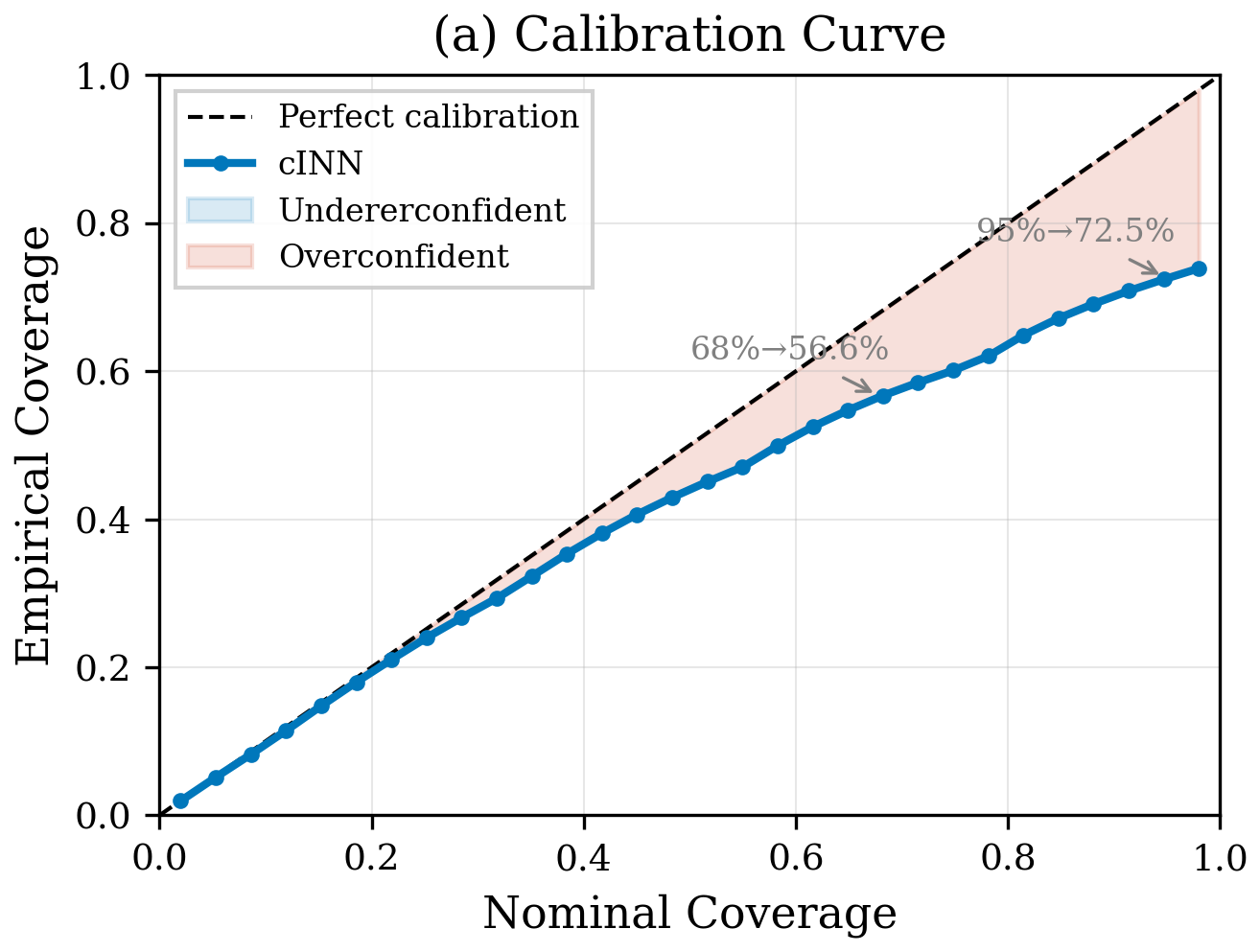}
\caption{Calibration curve of the open-loop predictions. The model
is mildly overconfident at high nominal levels.}
\label{fig:calibration}
\end{figure}

Fig.~\ref{fig:calibration} shows the calibration curve, comparing
the nominal coverage (the probability mass the model places inside
a central predictive interval) against the empirical coverage (the
fraction of test samples that fall inside it). A well-calibrated
model lies on the diagonal. The cINN stays systematically below it
on the upper half, reaching $57.7\,\%$ at $68\,\%$ nominal and
$73.9\,\%$ at $95\,\%$ nominal---mild overconfidence rather than
underconfidence; mean CRPS is $0.0915$ (post-hoc temperature
$T = 1.214$). Log-probability and absolute prediction error
correlate at Pearson $\rho = -0.60$: low-confidence samples are
those the model tends to get wrong. The flow stays numerically
invertible (round-trip reconstruction mean error
$2.8\!\times\!10^{-5}$ in standardised units).

\subsection{Closed-Loop Deployment}

\begin{figure}
\centering
\includegraphics[width=\columnwidth]{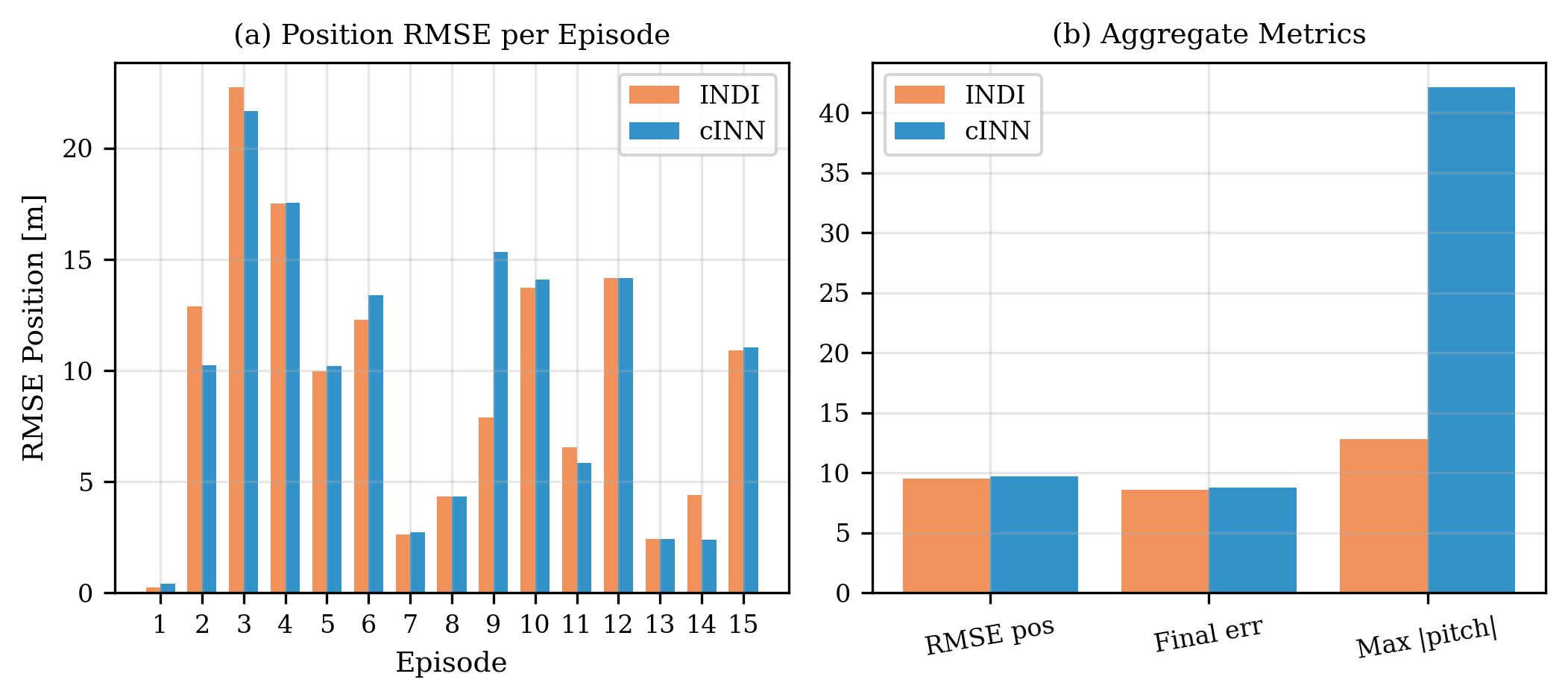}
\caption{Closed-loop comparison over $15$ scenarios.
(a)~Position RMSE per episode. (b)~Aggregate metrics.}
\label{fig:cl_summary}
\end{figure}

Across $15$ held-out scenarios the cINN reaches mean position RMSE
$9.7\,\mathrm{m}$ (INDI: $9.5\,\mathrm{m}$) and final position
error $8.8\,\mathrm{m}$ (INDI: $8.7\,\mathrm{m}$). The aggregate
peak pitch of $42^\circ$ (cINN) vs.\ $13^\circ$ (INDI) conflates
two regimes: in successful scenarios the cINN stays inside the
INDI envelope, while failure scenarios trigger the simulator's
attitude-safety break and dominate the maximum. The per-episode
distribution (Fig.~\ref{fig:cl_summary}) is bimodal. We sort the
scenarios into four tiers: \textit{success} ($\{4, 5, 8\}$),
\textit{acceptable} ($\{7, 10, 12, 15\}$), \textit{poor}
($\{3, 6, 9, 13\}$), and \textit{failure} ($\{1, 2, 11, 14\}$);
$7/15 \approx 47\,\%$ are success or acceptable.

\subsection{Success Cases}
\label{subsec:success}

In the successful and acceptable scenarios, the cINN remains within the
same qualitative operating envelope as the INDI baseline. These cases are
characterised by slowly varying velocity references and pitch excursions
that stay below approximately $\pm 20^\circ$. In this regime, which is
well covered by the training distribution, the cINN follows the INDI
trajectory with comparable settling behaviour and without triggering the
simulator's attitude-safety break.

The success cases also reveal a consistent difference between the learned
controller and the teacher. While the mean motor command remains close to
the INDI command level, the cINN output contains more high-frequency
content. This additional variation does not destabilise the plant in the
successful scenarios, but it indicates that the learned inverse-dynamics
model does not reproduce the filtering and smoothness properties of the
teacher perfectly.

\subsection{Failure Modes}
\label{subsec:failures}

The four failure scenarios separate into two distinct mechanisms.

\begin{figure}
\centering
\includegraphics[width=\columnwidth]{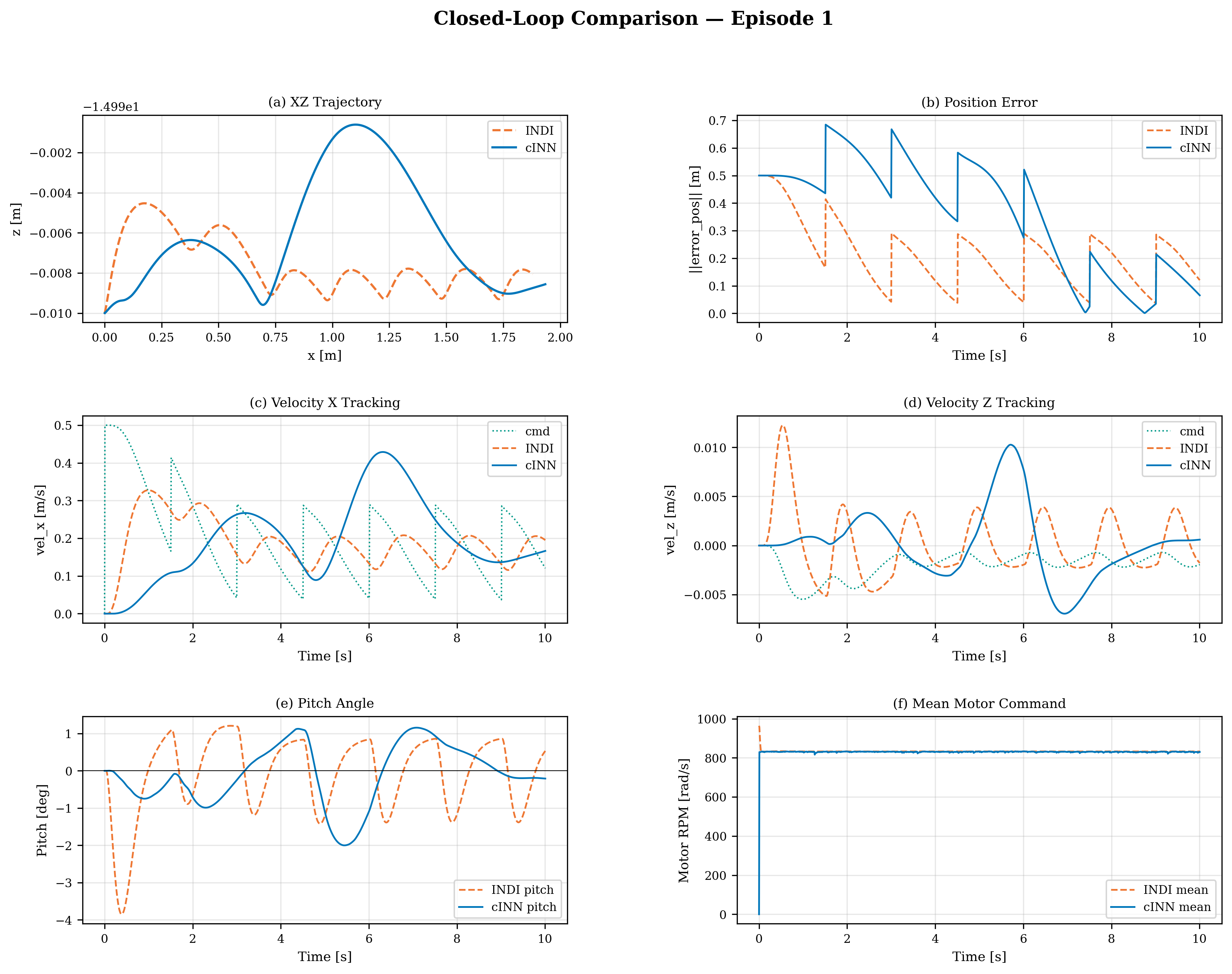}
\caption{Closed-loop response for Scenario~$1$ (Tier~D, phase-lag
failure).}
\label{fig:cl_scenario1}
\end{figure}

\emph{Command-tracking phase lag} (Scenario~$1$,
Fig.~\ref{fig:cl_scenario1}): the velocity command alternates
between $+0.5$ and $-0.5\,\mathrm{m/s}$ at roughly
$1.5\,\mathrm{s}$ intervals. The cINN remains stable over the full
$12.5\,\mathrm{s}$ episode---peak pitch below $3^\circ$ and mean
motor command at the INDI level---but the velocity response
(c,~d) trails INDI by about one switching period, and position
error~(b) accumulates as an oscillation rather than diverging. The
lag follows from the conditioning structure: $c_t$ encodes the
instantaneous error but not its rate of change, so the model
cannot anticipate a direction reversal.

\begin{figure}
\centering
\begin{minipage}[t]{0.48\columnwidth}\centering
  \includegraphics[width=\linewidth]{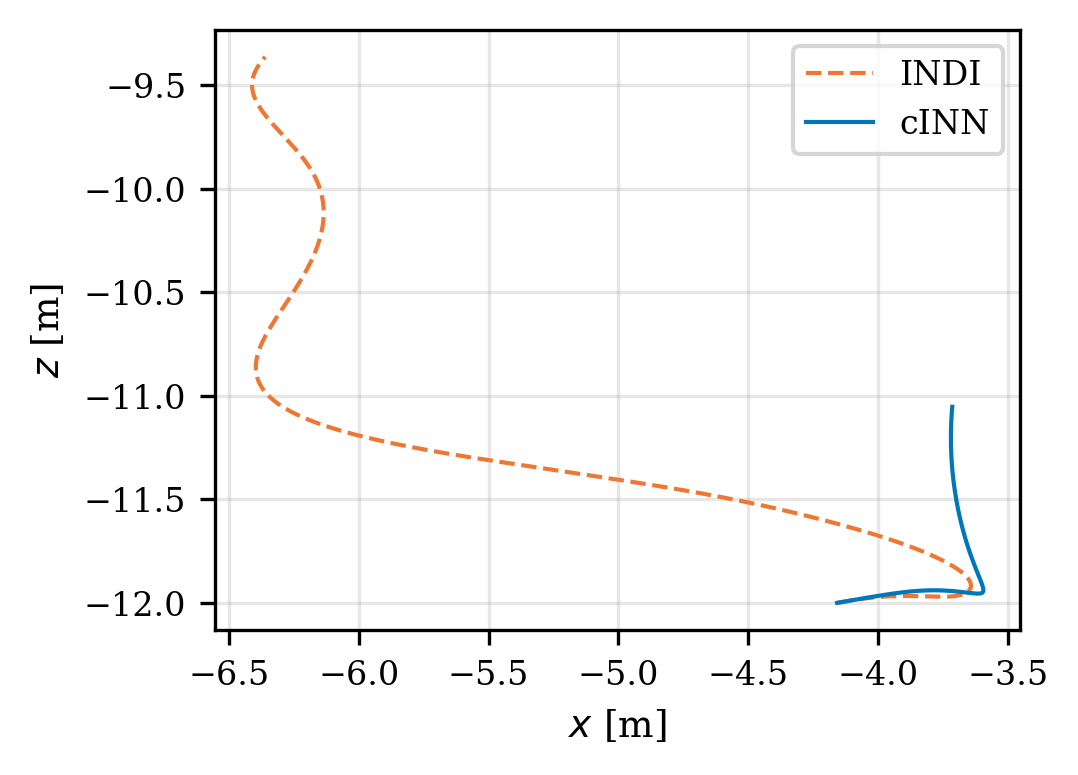}\\
  (a) trajectory
\end{minipage}\hfill
\begin{minipage}[t]{0.48\columnwidth}\centering
  \includegraphics[width=\linewidth]{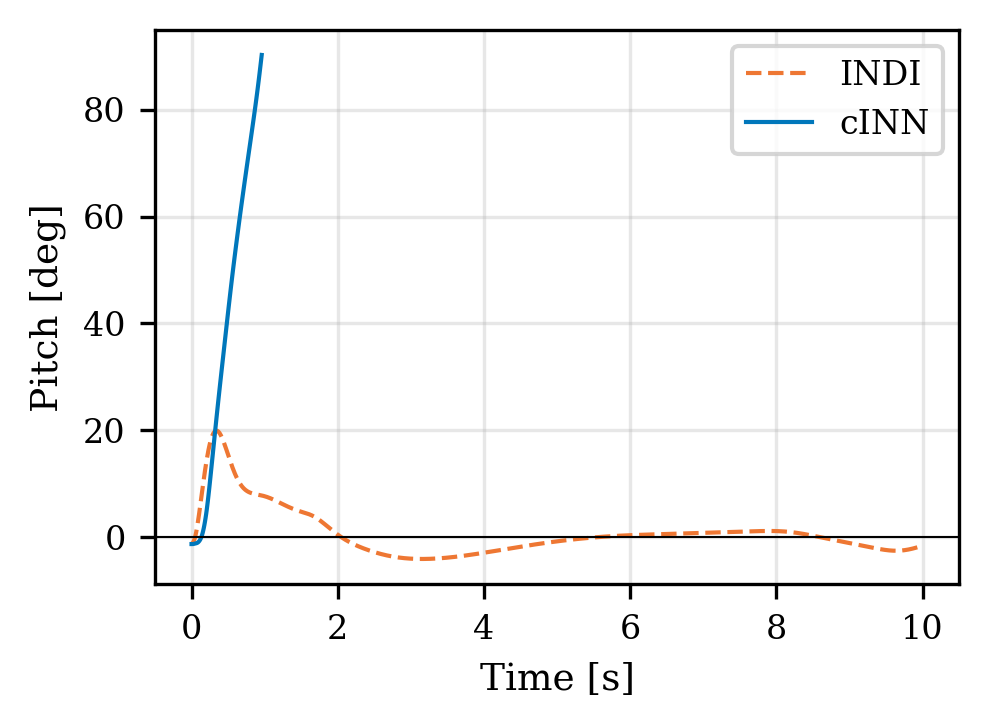}\\
  (b) pitch
\end{minipage}
\caption{Attitude divergence in Scenario~$11$. The cINN reaches
$90^\circ$ pitch within $1\,\mathrm{s}$; INDI completes the same
reference stably.}
\label{fig:cl_crash}
\end{figure}

\emph{Attitude divergence} (Scenarios~$2$, $11$, $14$,
Fig.~\ref{fig:cl_crash}): within the first two seconds an
aggressive velocity step starting from non-zero pitch drives the
cINN to a large motor differential, the pitch crosses
$\pm 30^\circ$, and the simulator's attitude safety break
terminates the episode. INDI handles the same references stably,
so the failure is specific to the learned controller. The common
feature is a rapid command change from a non-zero attitude---a
combination underrepresented in the training distribution, and the
regime in which the predictive log-probability is lowest,
consistent with the $\rho = -0.60$ correlation above.

\section{Conclusion}
\label{sec:conclusion}

We trained a conditional invertible neural network as a probabilistic
inverse-dynamics model for a 2-D X8 coaxial multicopter and evaluated it
against an INDI teacher in open- and closed-loop. Open-loop, the cINN
reproduces the teacher's eight-channel motor commands at $R^2 = 0.944$.
Closed-loop, it reaches a comparable mean position RMSE to INDI over
$15$ held-out scenarios, but with bimodal behaviour: $47\,\%$ of
scenarios are tracked successfully or acceptably, while the remaining
cases expose clear failure modes.

The probabilistic output is informative but not fully calibrated:
empirical coverage remains below the nominal levels at high coverage,
indicating mildly overconfident predictive intervals. Nevertheless,
log-probability correlates with absolute prediction error at
$\rho=-0.60$, giving a useful runtime confidence signal. The two observed
failure modes suggest direct extensions: temporal context or command
derivatives to reduce phase lag, and targeted data generation for
aggressive commands from non-zero attitude. A hybrid controller, in which
the cINN acts only above a confidence threshold and INDI takes over
otherwise, is the most immediate route towards safer deployment. Full
6-DoF flight and hardware deployment further require onboard state
estimation and reducing the current $47.5\,\mathrm{ms}$ inference time.

\section*{Declaration of Generative AI}
The authors used DeepL Write for language editing and Claude (Anthropic)
for code-related assistance. All content was reviewed and edited by the
authors, who take full responsibility for the publication.

\bibliography{references}








\end{document}